\PassOptionsToPackage{table,dvipsnames}{xcolor}

\documentclass{article}

\usepackage[preprint]{arxiv}

\usepackage[utf8]{inputenc} 
\usepackage[T1]{fontenc}    
\usepackage{hyperref}       
\usepackage{url}            
\usepackage{booktabs}       
\usepackage{amsfonts}       
\usepackage{nicefrac}       
\usepackage{microtype}      
\usepackage{xcolor}         

\usepackage{graphicx}
\usepackage{booktabs}
\usepackage{tabularx}
\usepackage{textcomp}
\usepackage{multirow}
\usepackage{bbm}
\usepackage{tikz}
\usepackage{comment}
\usepackage{graphicx}
\usepackage{amsmath}
\usepackage{amssymb}
\usepackage{booktabs}
\usepackage{algorithm}
\usepackage{algpseudocode}
\usepackage[accsupp]{axessibility} 
\usepackage{xfrac}
\usepackage{color}
\usepackage{wrapfig}
\usepackage{microtype}
\usepackage{tabularx}
\usepackage{booktabs}
\usepackage{multirow}
\usepackage{blindtext}
\usepackage{xspace}
\usepackage{pifont}

\usepackage{hyperref}
\usepackage[capitalize]{cleveref}
\crefname{section}{Sec.}{Secs.}
\crefname{algorithm}{Alg.}{Algs.}
\Crefname{section}{Section}{Sections}
\Crefname{table}{Table}{Tables}
\crefname{table}{Tab.}{Tabs.}

\newcolumntype{L}[1]{>{\raggedright\let\newline\\\arraybackslash\hspace{0pt}}m{#1}}
\newcolumntype{C}[1]{>{\centering\let\newline\\\arraybackslash\hspace{0pt}}m{#1}}
\newcolumntype{R}[1]{>{\raggedleft\let\newline\\\arraybackslash\hspace{0pt}}m{#1}}

\newcommand{\method}{EgoForce}

\usepackage[accsupp]{axessibility}  
\usepackage{hyperref}

\title{\method: Robust Online Egocentric Motion Reconstruction via Diffusion Forcing}

\author{Inwoo Hwang\textsuperscript{1},\quad Donggeun Lim\textsuperscript{1},\quad Hojun Jang\textsuperscript{1},\quad Young Min Kim\textsuperscript{1$\dagger$}  \\ \\ 
\textsuperscript{1} Seoul National University \\ \\ 
{\tt \href{https://inwoohwang.me/EgoForce}{https://inwoohwang.me/EgoForce}}
}

\begin{document}

\maketitle

\begin{abstract}
With recent advances in embodied agents and AR devices, egocentric observations are readily available as input for real-world interactive online applications.
However, egocentric viewpoints can only sporadically observe hands, in addition to the estimated head trajectory. 
We propose \textbf{\method}, an online framework for reconstructing long-term full-body motion from noisy egocentric input.
While existing generative frameworks can robustly handle noisy and sparse measurements, they assume a fixed-length observation window is available and are thus not suitable for real-time applications. 
Faster inference often relies on autoregressive prediction, sacrificing robustness.
In contrast, we adopt a diffusion-based method with a temporally asymmetric noise schedule inspired by Diffusion Forcing. 
Specifically, our approach models temporally evolving uncertainty and incrementally denoises states as new streaming observations arrive. 
Combined with a noise-robust imputation strategy, \method~progressively generates stable and coherent full-body motion under strict causal constraints.
Experiments demonstrate that our online framework outperforms existing online and offline methods, enabling long-horizon, full-body motion reconstruction in challenging egocentric scenarios.

\end{abstract}
\section{Introduction}
\label{sec:1_introduction}

As an increasing number of datasets capture diverse daily interactions from AR glasses or embodied agents~\cite{egobody, grauman2024egoexo4d, patel2025uniegomotion, ma24eccv, banerjee2024hot3d, fan2023arctic, damen2022rescaling, khirodkar2023egohumansegocentric3dmultihuman}, accurate reconstruction of human motion can accelerate advances in applications such as augmented and virtual reality, daily-life assistance, and real-time human interaction~\cite{sui2025surveyhumaninteractionmotion}.
Such dataset typically contains the locations of the head-mounted devices and, when interacting hands are within the limited field of view, their locations as well, resulting in sparse and noisy body-part observations.
Additionally, to fully leverage real-time data in interactive applications, the reconstruction should progressively track motions over a long time horizon without access to future observations.
These sparse, noisy, and strictly causal setups pose unique challenges that fundamentally differ from offline motion recovery, which assumes access to a clean, complete temporal context.

Existing approaches struggle to satisfy competing requirements. Online models that satisfy causality are often deterministic~\cite{barquero2025rolling, avatarjlm, jiang2024egoposer}, making them fragile under sparse or corrupted observations and limiting their overall modeling capacity. Conversely, recent diffusion-based generative models have proven effective at modeling uncertainty and recovering missing signals~\cite{yi2025egoallo, cho2025hamos, patel2025uniegomotion, guzov_jiang2025hmd2}; however, they typically rely on offline sequence-level denoising or autoregressive window-level sampling. 
In these scenarios, the denoising process recovers motion within a fixed temporal window, either offline or through autoregressive window-level sampling, rather than progressively refining a persistent per-frame streaming state, which is incompatible with streaming or frame-level online applications.
This leaves a critical mismatch between the modeling capacity required for robust egocentric motion reconstruction and the strict latency constraints of online systems.

To bridge this gap, we formulate online egocentric motion reconstruction as a causal generation problem. 
As the generative model processes the motion sequence within a temporally shifted window at each time step, the motion reconstruction is conditioned only on the past and current egocentric observations.
In the absence of future observations, unseen motion is estimated stochastically rather than deterministically, with uncertainty increasing monotonically with temporal distance from the present.
As new observations become available, the model traverses the temporal window and updates its uncertainty estimates accordingly, thereby refining its predicted motion to smoothly connect with past motion and respect the current egocentric measurements.

Our formulation preserves the expressive power of diffusion models while making them compatible with online inference.
We instantiate this formulation inspired by Diffusion Forcing~\cite{chen2024diffusionforcingnexttokenprediction}, which controls the denoising schedule of diffusion-based generative models to process flexible constraints.
Under strict causal constraints, we adapt temporally asymmetric frame-wise noise levels in which past motion remains fixed, the present is partially observed, and uncertainty increases toward the future.
The denoising process incrementally refines the current prediction using a small number of denoising steps at each time step as new streaming egocentric observations become available.
In addition, we incorporate a noise-robust imputation strategy that anchors reliable observations while allowing the model to infer plausible motion for unobserved joints.
Together, these components enable stable and coherent full-body motion reconstruction from sparse and corrupted egocentric inputs in an online manner.

\begin{figure*}[t]
\begin{center}
\includegraphics[width=0.9\linewidth]{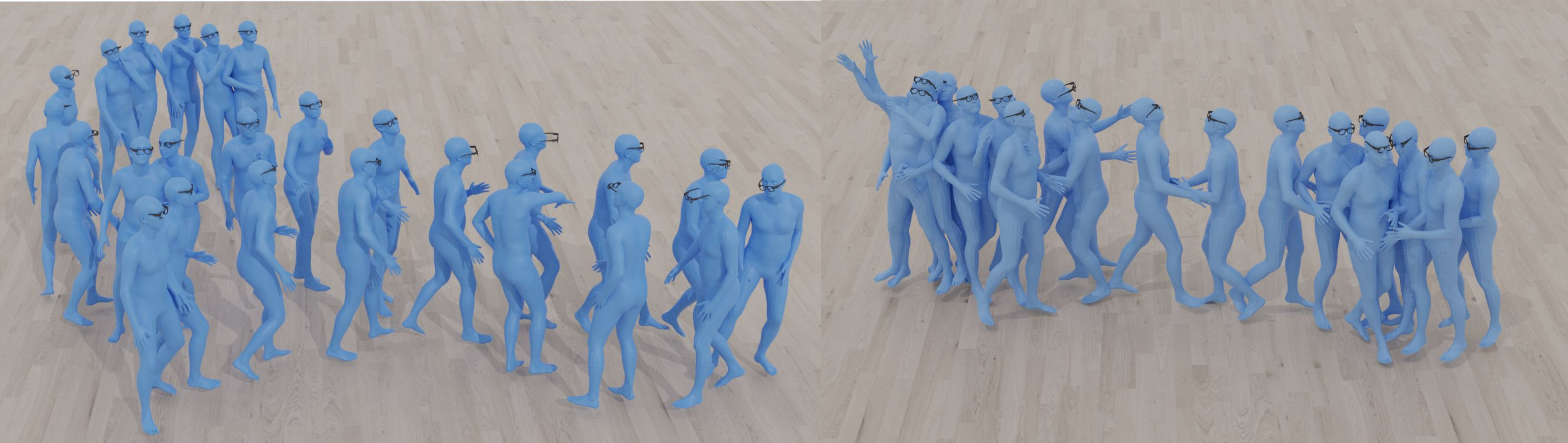}
\vspace{-0.5em}
\caption{
    \method~reconstructs full-body motion over long sequences in a strictly online manner. By progressively refining future predictions using both incoming observations and the generated motion history, it preserves temporal coherence and effectively mitigates long-term drift. We visualize two motion sequences, consisting of 610 and 660 frames, respectively.
}
\label{fig:long}
\end{center}
\vspace{-2em}
\end{figure*}

Our main contributions can be summarized as follows:
1) We formulate online egocentric motion reconstruction as a causal generation problem with temporally evolving uncertainty.
2) We adapt diffusion-based generative modeling to the online setting by restricting denoising to future motion states through Diffusion Forcing.
3) We demonstrate stable and robust full-body motion reconstruction from sparse and noisy egocentric signals under online constraints over long time horizons.
\section{Related Works}
\label{sec:2_related_works}

\subsection{Human Motion from Control Inputs}

Egocentric motion reconstruction can be framed as full-body motion reconstruction from sparse input signals.
Many previous works that generate motion from partial trajectories or keyframes assume offline access to complete and clean observations~\cite{wan2023tlcontrol, snapmogen2025, Pinyoanuntapong2025MaskControl, karunratanakul2023dno, Hwang_2025_CVPR, zhang2024rohm, karunratanakul2023gmd, motionlcm, meng2025absolute, guo2023momask}, such that models can exploit future information to maintain global coherence of motion.
Recent advancements enhance robustness against noisy or missing inputs~\cite{hwang2025motionsynthesissparseflexible, hwang2025scenemimotioninbetweeningmodeling} by adapting imputation techniques.
However, these formulations remain inapplicable to online interactive applications.
While RPM~\cite{barquero2025rolling} moves toward online reconstruction through a causal transformer, it remains deterministic and lacks the robustness of stochastic generative models.

\subsection{Egocentric Motion Reconstruction}

The development of head-mounted devices~\cite{ma24eccv} has introduced diverse datasets of egocentric measurements paired with full-body motions~\cite{egobody,patel2025uniegomotion,grauman2024egoexo4d}.
Prior approaches to egocentric motion reconstruction typically adopt causal sequence models, such as transformers~\cite{jiang2024egoposer, avatarjlm, jiang2022avatarposer}, to enable online applications. 
However, they are often deterministic and lack sufficient generative capacity to model the inherent uncertainty of human motion, whereas the egocentric observations are noisy and sparse input for full-body reconstruction~\cite{guzov_jiang2025hmd2, EgoLM}.
In contrast, diffusion-based generative models~\cite{li2023ego} have demonstrated strong robustness by incorporating a diffusion sampling process with guidance mechanisms~\cite{yi2025egoallo} or conditioning on egocentric images and control signals~\cite{patel2025uniegomotion}. Recent work further extends diffusion-based motion reconstruction to causal settings through autoregressive window-level inpainting~\cite{guzov_jiang2025hmd2}.
However, these diffusion-based methods either perform offline inference over the full observation window or rely on autoregressive window-level sampling, whereas our goal is per-frame streaming reconstruction with incremental refinement as each new observation arrives.
To bridge this gap, EgoForce formulates online full-body motion reconstruction as a generative framework that enables progressive inference under sparse and noisy observations, combining the causality of online systems with the robust generative capacity of diffusion models.

\subsection{Causal Modeling with Diffusion Models}

Recent diffusion models~\cite{song2020denoising, ho2022classifierfreediffusionguidance} extend their capability to fulfill the requirements of streaming content generation~\cite{MotionStreamer, maluleke2025diffusionforcingmultiagentinteraction, camdm, shi2024amdm, Zhao:DartControl:2025, yu2026causal} and online decision making~\cite{wu2025uniphys, truong2024pdp, chi2023diffusionpolicy}.
TEDi~\cite{zhang2024tedi} and Rolling Diffusion Models~\cite{ruhe2024rollingdiffusionmodels} employ a rolling-window formulation, in which diffusion models generate sequences by repeatedly denoising a short temporal window.
More recent efforts reformulate diffusion as a strictly causal or autoregressive process by introducing frame-wise varying noise levels and causal sampling schedules. 
Diffusion Forcing~\cite{chen2024diffusionforcingnexttokenprediction} assigns independent noise levels to individual frames during training, enabling strict causal generation while preserving the expressive power of diffusion models. SDP~\cite{hoeg2024streamingdiffusionpolicyfast} applies a streaming diffusion framework to robotic policy learning, enabling online action generation. 
Autoregressive diffusion variants~\cite{ardiff} further factorize denoising across time steps to support online inference. 
Our work extends this line of research to online egocentric motion reconstruction.
By explicitly modeling the temporal evolution of uncertainty online, EgoForce effectively preserves the expressive capacity of generative diffusion while satisfying the strict causal constraints of real-world egocentric systems.
\section{Method}
\label{sec:3_method}

We present an online, causal framework for reconstructing full-body human motion from sparse egocentric observations.
We first formulate the problem setting and causal constraints (\Cref{sec:formulation}),
then describe the training procedure based on frame-wise noise corruption under causal conditioning (\Cref{sec:training}),
and finally introduce the progressive denoising strategy for causal online inference (\Cref{sec:sampling}).

\subsection{Problem Formulation}
\label{sec:formulation}

We consider an \textbf{online egocentric setting} where an image stream $\mathcal{I} = \{\mathbf{I}_1, \dots, \mathbf{I}_n\}$ is captured by a head-mounted device. From each image $\mathbf{I}_t$, we extract a sparse set of kinematic control signals. Specifically, at time step $t$, the kinematic control signal $\mathbf{c}_t$ is defined as:
\begin{equation}
\mathbf{c}_t = \left(\mathbf{h}_t, (\mathbf{w}^l_t, \mathbf{v}^l_t), (\mathbf{w}^r_t, \mathbf{v}^r_t) \right),
\label{eq:input_condition}
\end{equation}
where $\mathbf{h}_t \in SE(3)$ denotes the global head pose, represented by translation and a continuous 6D rotation representation, and $\mathbf{w}^{l/r}_t \in SE(3)$ represents the left and right wrist poses estimated from $\mathbf{I}_t$ or provided as proxy sparse controls depending on the benchmark. To account for frequent self-occlusions and the limited field-of-view of egocentric cameras, we incorporate visibility indicators $\mathbf{v}^{l/r}_t \in \{0, 1\}$. Here, $\mathbf{v}^{l/r}_t = 1$ indicates that the corresponding wrist pose is reliably observed, while $\mathbf{v}^{l/r}_t = 0$ denotes missing data.

Our goal is to learn a causal generative model $\mathcal{G}$ that reconstructs the current body pose $\mathbf{x}_t$ in an online manner where each pose $\mathbf{x}_t \in \mathbb{R}^d$ represents the full-body configuration at time step $t$ with a \textbf{strict causal constraint}: the prediction of $\mathbf{x}_t$ must depend solely on the history of previously generated motions $\mathbf{x}_{<t}$ and the observed control signals up to the current time step, $\mathbf{c}_{\leq t}$ and $\mathbf{I}_{\leq t}$. 
Accordingly, we formulate the generation process by modeling the conditional distribution of the current pose $\mathbf{x}_t$ given the past motion context and available observations
\begin{equation}
p(\mathbf{x}_t \mid \mathbf{x}_{<t}, \mathbf{c}_{\leq t}, \mathbf{I}_{\leq t}).
\label{eq:posterior}
\end{equation}
This causal formulation is designed to enable online motion reconstruction, while the probabilistic formulation maintains robustness against the sparse and intermittent nature of egocentric input.

\subsubsection{Motion Representation}

We represent full-body motion in a canonical coordinate system centered on the head-mounted device (e.g., AR glasses)~\cite{yi2025egoallo, patel2025uniegomotion}, with the $+z$ axis aligned with its forward direction. 
Each pose $\mathbf{x}_t$ includes a central reference point corresponding to the device and 22 body joints, all described by global translation and continuous 6D rotation representation~\cite{hwang2025scenemimotioninbetweeningmodeling}. Foot contact indicators for both feet are also incorporated.

\subsection{Training Pipeline with Frame-wise Noise Corruption under Causal Conditioning}
\label{sec:training}

We encode the temporal context of motion by observing a sequence of full-body 3D motion within the sliding window of length $h+1+f$.
Centered at the current time step $t$, the temporal window contains motions of a fixed history length $h$ and a future prediction horizon $f$ with respect to the current time step $t$ as $\mathbf{X}_t = \{\mathbf{x}_{t-h}^{k_{t-h}}, \dots, \mathbf{x}_{t+f}^{k_{t+f}}\}$.
Note that we are using the subscript $t$ to denote the temporal index of motion frames, while we use the superscript $k \in \{0, \ldots, K\}$ ($K$ denotes the maximum number of diffusion steps) to denote the diffusion time step where the denoising process $\mathbf{x}_t^K \rightarrow \mathbf{x}_t^0$ recovers the clean sequence $\mathbf{x}_t^0$ from a noisy one $\mathbf{x}_t^K$ with $K$ maximum number of diffusion steps.

\begin{figure*}[t]
\begin{center}
\includegraphics[width=0.9\linewidth]{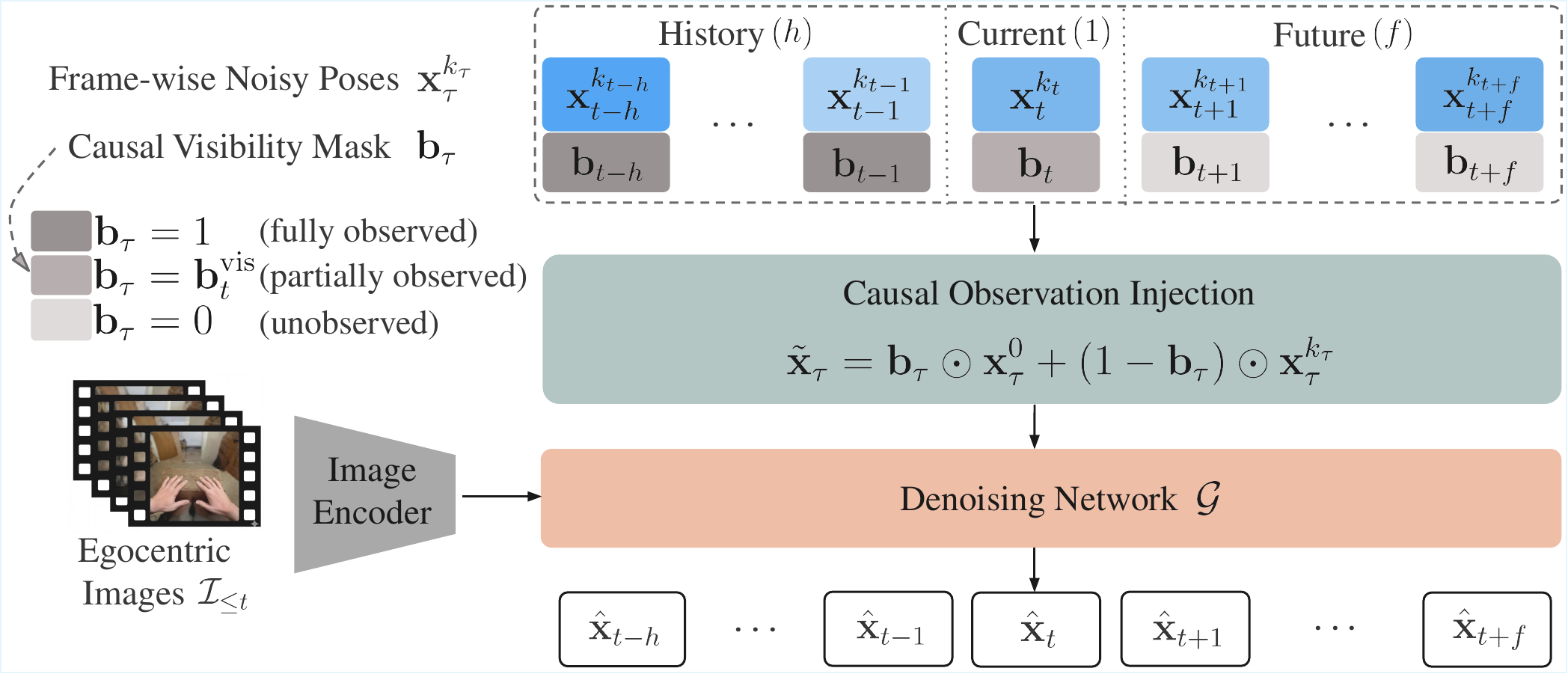}

\caption{
    \textbf{Training pipeline with frame-wise noise corruption under causal conditioning.} A motion segment centered at time step $t$ is corrupted with heterogeneous diffusion noise $k_\tau$ across frames. Egocentric causal observations are injected, and the denoising network $\mathcal{G}$ is trained to reconstruct the clean motion sequence conditioned on causal egocentric context. 
}
\label{fig:training}
\end{center}
\vspace{-1.5em}
\end{figure*}

To train the online framework, we design the frame-wise noise level $k_\tau$ of the temporal window to introduce noisy samples for the diffusion model $\mathbf{x}_\tau$ paired with visibility mask $\mathbf{b}_\tau$, which are composed to serve as input to the denoising network $\mathcal{G}$, as illustrated in Figure~\ref{fig:training}.

\subsubsection{Diffusion Model with Frame-wise Noise}

Unlike standard diffusion training, which applies a uniform noise level across all frames in a sequence, we assign independent, varying per-frame noise levels, adopting the Diffusion Forcing paradigm~\cite{chen2024diffusionforcingnexttokenprediction}.
For each training step, we sample a ground-truth motion window $\mathbf{X}_t^0 = \{\mathbf{x}_{t-h}^0, \dots, \mathbf{x}_{t+f}^0\}$ centered around the current time step $t$. 
We define a vector of diffusion noise levels over the temporal window,
$
\mathbf{k}_t = \{k_{t-h}, \dots, k_{t+f}\},
$
where each element is independently sampled as
\begin{equation}
k_\tau \sim \mathcal{U}\{0, 1, \dots, K\}, \quad \tau \in [t-h, t+f].
\label{eq:noise_levels}
\end{equation}
Each ground-truth pose $\mathbf{x}_\tau$ is then perturbed according to its assigned noise level $k_\tau$, producing the noisy pose $\mathbf{x}_\tau^{k_\tau}$.
This approach enables the model to denoise motion trajectories under heterogeneous, frame-wise uncertainty, facilitating flexible noise scheduling during inference.

\subsubsection{Causal Observation Injection}
\label{sec:causal_injection}

To strictly enforce online causality during training, we inject sparse body-part observations using a causal visibility mask $\mathbf{b}_\tau \in \{0,1\}^d$, as illustrated in Figure~\ref{fig:training}. The mask follows the online information structure: past frames are fully observable from generated full-body history, the current frame is partially observable through the head and visible wrists, and future frames are unobserved. For the current frame, the active observations are determined by sparse egocentric signals, taking into account the wrist visibility indicators $\mathbf{v}_t$, while all unobserved joints remain masked.

We construct the causally injected pose
\begin{equation} \label{eq:causal_imputation}
\tilde{\mathbf{x}}_\tau = \mathbf{b}_\tau \odot \mathbf{x}^0_\tau + (1 - \mathbf{b}_\tau) \odot \mathbf{x}^{k_\tau}_\tau,
\end{equation}
which anchors reliable observations while preserving stochastic uncertainty elsewhere. This operation enforces consistency with reconstructed past motions and current sensor input, while allowing the model to predict plausible future trajectories.

To explicitly inform the network which dimensions are reliable, we concatenate the injected pose with its visibility mask, $\bar{\mathbf{x}}_\tau = [\tilde{\mathbf{x}}_\tau, \mathbf{b}_\tau]$, and construct the augmented motion window $\bar{\mathbf{X}}_t = \{\bar{\mathbf{x}}_{t-h}, \dots, \bar{\mathbf{x}}_{t+f}\}$.

\subsubsection{Noise-Robust Causal Observation Injection}
\label{sec:noisy_causal_injection}

To improve robustness to noisy egocentric inputs, we synthetically corrupt the egocentric control signals $\mathbf{c}_t$ into noisy versions $\mathbf{c}_t^\text{noisy}$ during training and train a denoiser using the corresponding clean motion data $\mathbf{x}_t$, following~\cite{hwang2025scenemimotioninbetweeningmodeling}. 
Since the model prediction is unstable with a large number of diffusion steps $k \geq K^*$, a noisy observation signal is applied only during the early diffusion steps up to a threshold $K^*$, whereas later denoising stages rely solely on diffusion model predictions. 
The causally injected pose is defined as
\begin{equation}
\tilde{\mathbf{x}}_\tau =
\begin{cases}
\mathbf{b}_\tau \odot \mathbf{x}^0_\tau(\mathbf{c}_t^{\mathrm{noisy}})
+ (1 - \mathbf{b}_\tau) \odot \mathbf{x}^k_\tau,
& k \ge K^* \\
\mathbf{x}^k_\tau,
& k < K^*.
\end{cases}
\end{equation}

\subsubsection{Conditioning and Objective}

In addition to the sparse kinematic observations $\mathbf{c}_t$ that are injected directly into the motion representation, the causal egocentric image stream
\[
\mathbf{I}_{\le t} = \{\mathbf{I}_{t-h}, \dots, \mathbf{I}_t\}
\]
is encoded by an image encoder~\cite{oquab2023dinov2} and fused into the denoising network $\mathcal{G}$ via cross-attention. This visual context resolves ambiguities caused by missing joints and provides semantic cues for motion reconstruction.

The model is trained to directly predict the ground truth motion window $\mathbf{X}^0_t$ by minimizing the expected $L_2$ reconstruction error
\begin{equation}
\mathcal{L} = \mathbb{E}_{\mathbf{X}^0_t, \mathbf{k}} \left[ \left\| \mathcal{G}(\bar{\mathbf{X}}_t, \mathbf{k},  \mathbf{I}_{\le t}) - \mathbf{X}^0_t \right\|^2_2 \right].
\end{equation}

\subsection{Causal Online Inference with Progressive Denoising Refinement}
\label{sec:sampling}

\begin{figure*}[t]
\begin{center}
\includegraphics[width=0.85\linewidth]{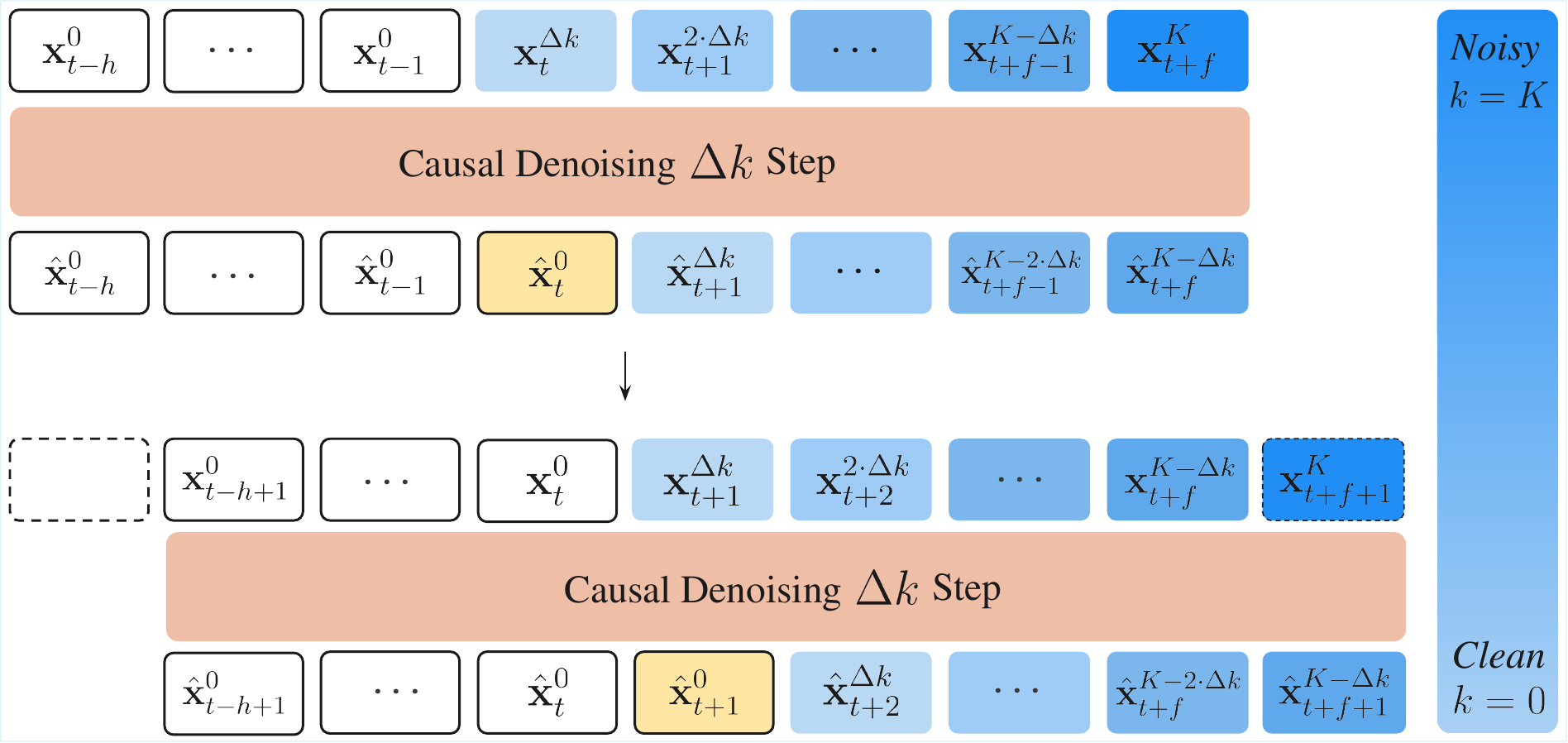}
\caption{
    \textbf{Causal online inference with progressive denoising refinement.} At each time step, the temporal window is shifted forward to reuse previously denoised states as warm-starts, while a new future frame is initialized with Gaussian noise. Causal egocentric observations are injected, and the denoising network performs a fixed $\Delta k$ refinement step to fully denoise the current pose while progressively refining future predictions under increasing uncertainty. as detailed in the Appendix.
}
\label{fig:sampling}
\end{center}
\vspace{-1em}
\end{figure*}

After the model is trained to denoise signals under heterogeneous noise conditions, we leverage its behavior and design the inference step for online generation.
The inference process is formulated as a \textbf{progressive denoising refinement} procedure, with a carefully designed per-frame noise level.
The temporal sequence $\mathbf{x}_t \in \mathbf{X}_t$ ensures coherence to the previous generation $\hat{\mathbf{x}}_{t-h:t-1}$, while respecting the current, and possibly noisy, egocentric observations $(\mathbf{c}_t, \mathbf{I}_t)$. Simultaneously, it anticipates future motion trajectories within a consistent prediction horizon $f$.
The process is illustrated in Figure~\ref{fig:sampling} and also described below.

\subsubsection{Initializing the Sequence}
At the very first step of inference ($t=0$), the system initializes the entire buffer with Gaussian noise. 
We denoise the initial sequence to recover a stable trajectory sample.
During this stage, the historical frames ($h$) and the current frame are fully denoised ($k=0$), while the future horizon ($f$) is partially denoised to follow the predefined noise for the future horizon as follows.

\paragraph{Noise Scheduling for Future Horizon.}

To capture the increasing uncertainty across the future prediction horizon $f$, we assign a noise schedule that increases monotonically with temporal distance. For a frame at a relative offset $i \in \{0, \dots, f\}$ from the current step $t$, the target noise level $k_{t+i}$ is defined as
\begin{equation}
k_{t+i} = i \cdot \Delta k, \quad \Delta k = \frac{K}{f+1}.
\label{eq:future_noise}
\end{equation}
This temporal scheduling ensures the current frame $\mathbf{x}_t$ reaches a clean state ($k_t = 0$), while future predictions are increasingly noisy states, with the furthest terminal frame $\mathbf{x}_{t+f}$ approaching the maximum diffusion step $K$.

\subsubsection{Online Inference Pipeline}

As time advances from $t$ to $t+1$, the observed motion sequence $\mathbf{X}_t$ is shifted by a single time step, sharing $h+f$ temporal interval.
The inference loop consists of three stages below, which is visualized in Figure~\ref{fig:sampling}.

1. \textit{Temporal Shift and Warm-start.} As time evolves, the temporal window slides forward by one frame. The oldest history frame $\mathbf{x}_{t-h}$ is discarded, and all latents in the buffer are shifted to the left. These shifted frames serve as high-quality warm-starts because they retain the denoising progress from the previous time step. To maintain the horizon, a new terminal frame $\mathbf{x}_{t+f+1}$ is introduced and initialized with standard normal noise: $\mathbf{x}_{t+f+1}^{K} \sim \mathcal{N}(\mathbf{0}, \mathbf{I})$.

2. \textit{Causal Condition Injection.} Upon receiving new egocentric observations at $t+1$, $(\mathbf{c}_{t+1}, \mathbf{I}_{t+1})$, we update the reliability masks $\mathbf{b}_\tau$ and inject the causal signals using the estimated signal $\hat{\mathbf{x}}_\tau$ instead of the ground truth motion $\mathbf{x}_\tau^0$ in Equation~(\ref{eq:causal_imputation}). Then, we construct the sequence $\bar{\mathbf{X}}_{t+1}$ augmented with the binary mask $\mathbf{b}_\tau$.

3. \textit{Progressive Denoising.} 
The denoising network $\mathcal{G}$ performs a fixed number of $\Delta k$ refinement steps. To mitigate compounding errors and distribution drift during long-horizon rollouts, we apply a stabilization trick by re-injecting a minimal noise level $n > 0$ into the historical latents $\mathbf{x}_{\tau \in [t-h+1, t]}$ before the denoising step
\begin{equation}
\mathbf{x}_\tau^{\max(k - \Delta k, 0)} \leftarrow \mathcal{G}(\text{InjectNoise}(\mathbf{x}_\tau^k, n), \bar{\mathbf{X}}_{t+1}, \mathcal{I}_{\le t+1}) \text{.}
\end{equation}
    
As we reuse the prediction from the shared period, our progressive refinement reverses only $\Delta k$ diffusion steps, rather than denoising the entire sequence from scratch.
Such a design significantly reduces the number of diffusion steps per progression, enabling real-time estimation of full-body motion.
We provide the formal algorithmic procedure in the appendix to facilitate a clearer understanding of our inference pipeline.
\section{Experiments}
\label{sec:4_experiments}

In this section, we present a comprehensive experimental evaluation of our method for online egocentric motion reconstruction. 
We first outline the evaluation setup (\Cref{sec:eval_details}).
Next, we compare our method with online and offline baselines for egocentric motion reconstruction (\Cref{sec:eval_ego}), and demonstrate that our approach achieves strong performance under strict online constraints. Finally, we evaluate the robustness of our approach under noisy input conditions (\Cref{sec:eval_robust}).

\subsection{Evaluation Details}
\label{sec:eval_details}
\paragraph{Dataset and Implementation Details.}

Our pipeline is implemented in PyTorch~\cite{pytorch} and trained on a single NVIDIA RTX 3090 GPU. We use the EE4D-Motion dataset~\cite{patel2025uniegomotion} as our primary benchmark, which is sampled at 10 FPS. During training, we randomly sample sliding windows consisting of $h + 1 + f$ frames, where $h$ and $f$ denote the history length and future lookahead horizon, respectively. In our default setting, we use $h = 5$ and $f = 19$, resulting in 25-frame windows.
We use $K = 100$ diffusion timesteps. Each causal denoising step spans $\Delta k = K / (f + 1) = 5$ diffusion steps. For stabilization sampling, we set the noise injection level to $n = 2$.
When training with noisy egocentric observations (\Cref{sec:eval_robust}), we inject zero-mean noise with a standard deviation into the input signals. The noise standard deviation is controlled by a noise level $l$, defined as $(l^{\circ}, l~\text{cm})$ for rotational and translational signals, respectively, following~\cite{hwang2025scenemimotioninbetweeningmodeling}. We set $l = 2$ and $K^* = 3$.
Additional details on the data processing and implementation are provided in the Appendix.

\paragraph{Baselines.}

We compare our method against several baselines, including online egocentric motion reconstruction models (AvatarJLM~\cite{avatarjlm} and RPM~\cite{barquero2025rolling}), an online diffusion-inpainting baseline (HMD$^2$~\cite{guzov_jiang2025hmd2}), and recent diffusion-based models (UniEgoMotion~\cite{patel2025uniegomotion} and EgoAllo~\cite{yi2025egoallo}).
For the offline methods, we train each model using 80 randomly sampled frames per window. During generation, the reconstructed windows are stitched together to produce the full motion sequence.
On the other hand, our online method processes the long-term trajectory progressively.
Evaluation metrics are computed over the full motion sequence. 
More details on baseline implementations are provided in the Appendix.

\paragraph{Evaluation Metrics.}
We evaluate our method across four dimensions: 
(i) Reconstruction Accuracy via MPJPE (m) and rotation error (MPJRE-F, Frobenius norm) for the full body, along with Head Position Error (Head PE) and Wrist Position Error (Wrist PE); 
(ii) Motion Quality using TMR-based~\cite{petrovich23tmr} semantic similarity and FID to assess realism; 
(iii) Motion Smoothness via Peak Jerk (PJ) and Area Under the Jerk (AUJ) to measure temporal stability; and 
(iv) Online Capability by reporting causality and per-frame inference latency. 
Detailed definitions and configurations are provided in the Appendix.

\subsection{Online Egocentric Motion Reconstruction}
\label{sec:eval_ego}
\begin{table*}[t]
\caption{Quantitative egocentric motion reconstruction results on the EE4D-Motion dataset, comparing online and offline baselines. Our method achieves state-of-the-art performance under strict online causal constraints. Best values are highlighted in \textcolor{blue}{blue} and second-best values in \textcolor{red}{red}. $^\dagger$ denotes our adapted re-implementation because the official code is not publicly available.}
    \centering
    \setlength{\tabcolsep}{6pt}
    \renewcommand{\arraystretch}{1.1}
    \scalebox{0.76}{
    \begin{tabular}{lcccccccccc}
    \toprule
    \multirow{2}{*}{Methods}
    & \multicolumn{4}{c}{Reconstruction $\downarrow$}
    & \multicolumn{2}{c}{Quality}
    & \multicolumn{2}{c}{Smoothness}
    & \multicolumn{2}{c}{Sampling}\\
    \cmidrule(lr){2-5}
    \cmidrule(lr){6-7}
    \cmidrule(lr){8-9}
    \cmidrule(lr){10-11}
     & Head PE & Wrist PE & MPJPE & MPJRE-F & Semantic $\uparrow$ & FID $\downarrow$ & PJ $\rightarrow$ & AUJ $\downarrow$ & Time~(s) $\downarrow$ & Online \\
    \midrule
    Real motions  & - & - & - & - & 0.999 & 0.001 & 0.329 & 10.7  & - & -  \\
    \midrule
    AvatarJLM~\cite{avatarjlm}  & - & 0.187 & 0.119 & 0.388 & 0.829 & 0.096 & 1.743 & 18.8 & {\color{red}{0.01}} & \ding{51}  \\
    RPM~\cite{barquero2025rolling}  & - & 0.108 & 0.115 & 0.364 & 0.907 & 0.042 & 0.453 & 11.1 & {\color{blue}{0.005}} & \ding{51}  \\
    HMD$^2$$^\dagger$~\cite{guzov_jiang2025hmd2} & - & 0.131 & 0.127 & 0.245 & 0.913 & 0.032 & 0.373 & 11.0 & 0.35 & \ding{51}   \\
    EgoAllo~\cite{yi2025egoallo}  & - & 0.079 & 0.168 & 0.612 & 0.869 & 0.045 & 0.912 & 14.3 & 84.9 & \ding{55}  \\
    UniEgoMotion~\cite{patel2025uniegomotion}  & {\color{blue}{0.034}} & 0.089 & 0.091 & 0.497 & 0.917 & 0.024 & 0.738 & 12.7 & 2.05 & \ding{55}  \\
    \method\;(Ours)  & 0.072 & {\color{red}{0.076}} & {\color{red}{0.073}} & {\color{red}{0.138}} & {\color{red}{0.987}} & {\color{red}{0.016}} & {\color{red}{0.318}} & {\color{red}{10.8}} & 0.08 & \ding{51} \\
    \midrule
    \;\; w/o image feat.  & 0.073 & 0.078 & 0.077 & 0.141 & 0.986 & 0.018 & 0.313 & 11.0 & 0.07 & \ding{51} \\
    \;\; w/o stabilize.  & 0.074 & 0.081 & 0.075 & 0.143 & 0.985 & 0.019 & 0.315 & 10.9 & 0.08 & \ding{51} \\
    \;\; $h=30$  & 0.138 & 0.117 & 0.129 & 0.271 & 0.933 & 0.086 & 0.298 & 12.9 & 0.08 & \ding{51} \\
    \;\; offline ver.  & {\color{red}{0.036}} & {\color{blue}{0.044}} & {\color{blue}{0.042}} & {\color{blue}{0.059}} & {\color{blue}{0.991}} & {\color{blue}{0.012}} & {\color{blue}{0.322}} & {\color{blue}{10.8}} & 3.19 & \ding{55} \\
    \bottomrule
    \end{tabular}
    }
    \label{tab:quantitative_eval_ego}
\end{table*}

\vspace{-1.0em}
\begin{figure*}[h!]
\begin{center}
\includegraphics[width=0.95\linewidth]{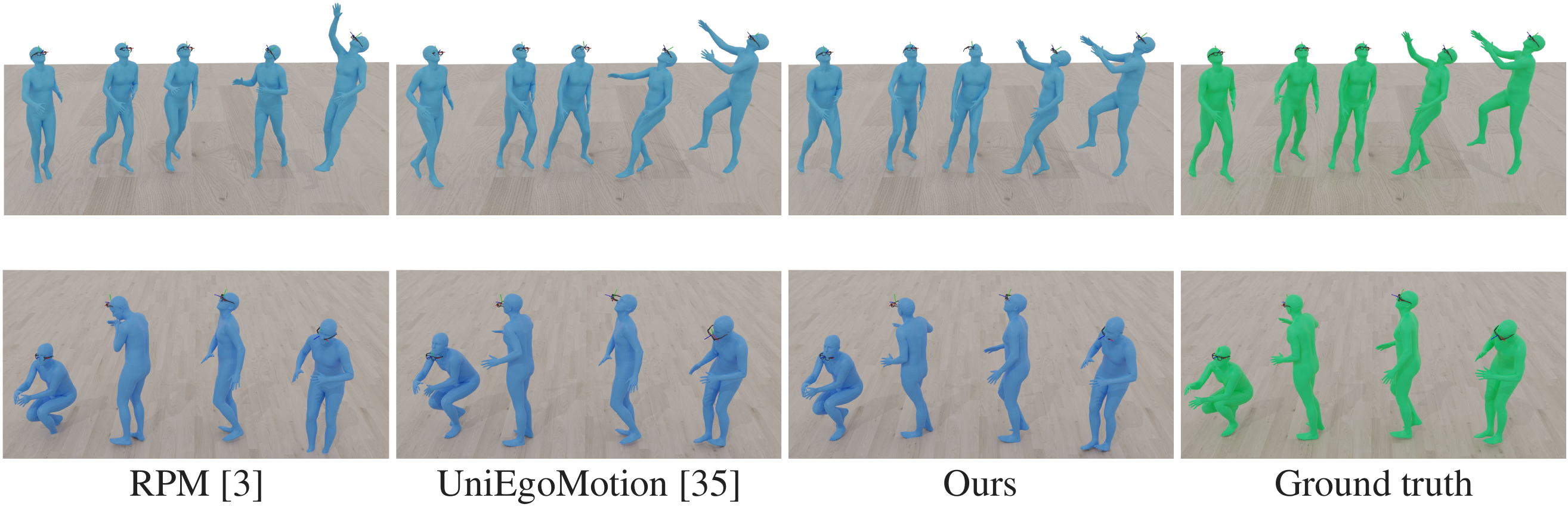}
\vspace{-0.0em}
\caption{
    Existing online methods (e.g., RPM~\cite{barquero2025rolling}) suffer from limited motion fidelity, whereas offline approaches (e.g., UniEgoMotion~\cite{patel2025uniegomotion}) rely on window-based generation and stitching, often leading to discontinuous motion at window boundaries. In contrast, our method generates globally coherent and smooth motion under strict causal constraints.
}
\label{fig:comparison}
\end{center}
\vspace{-1em}
\end{figure*}

Table~\ref{tab:quantitative_eval_ego} presents a quantitative comparison of our approach against various baseline configurations, including strong online and offline baselines. 
The supplementary video features the reconstructed motion sequences.

\paragraph{Baselines Analysis.} 
Existing online reconstruction methods satisfy strict causality and online constraints, but their limited modeling capacity often leads to inferior reconstruction accuracy and lower overall motion quality. In contrast, offline reconstruction methods achieve higher motion quality and lower reconstruction errors; however, they rely heavily on future control signals and extensive iterative denoising or additional guidance inference. This incurs substantial latency, making them unsuitable for online egocentric reconstruction settings. 

\paragraph{Reconstruction Accuracy and Motion Quality under Online Constraints.} 
Our method effectively bridges this gap between online capability and motion quality. 
Under strict online settings, it consistently outperforms existing online baselines by producing more accurate and temporally coherent full-body motions. Specifically, our approach achieves significantly lower FID, higher semantic similarity, and reduced reconstruction errors for both the full body and the head. Moreover, our approach remains competitive with offline diffusion-based methods while requiring substantially fewer inference steps and strictly operating without future information. These results demonstrate that our formulation enables both high-quality motion reconstruction and practical online deployment. 
In Figure~\ref{fig:comparison}, our results more faithfully follow the ground-truth motion while strictly adhering to causal online constraints.
The motion quality is better visualized in the supplementary video.

\paragraph{Long-Term Stability and Smoothness.} 
Furthermore, our approach exhibits exceptional stability in reconstructing long-term motion sequences. 
With future prediction conditioned on historical states, which is progressively refined with incoming egocentric observations,
our method effectively mitigates error accumulation over time. This continuous refinement leads to highly stable and smooth motion generation. In contrast to online approaches like RPM that attempt to enforce smoothness, or offline methods that typically generate fixed-length temporal windows and stitch them together (often resulting in discontinuous transitions at boundaries), our method ensures seamless and naturally smooth movements. This superior temporal stability is quantitatively validated by our Peak Jerk (PJ) and significantly lowered Area Under Jerk (AUJ) metrics. Figure~\ref{fig:long} presents representative long-horizon reconstructions, highlighting consistent global structure and smooth temporal transitions across extended motion sequences.

\paragraph{Ablation Study.}

We compare our model against several ablated variants to evaluate the impact of our specific design choices. By conditioning on image features, our model reconstructs more semantically similar motions, which is quantitatively verified by improved semantic similarity and motion quality scores. 
Furthermore, incorporating our stabilization trick improves reconstruction accuracy by reducing error accumulation during the sampling process. 
We additionally find that conditioning on a longer history context ($h=30$) accumulates errors and degrades performance; therefore, we limit the history length ($h$) to 5.

\paragraph{Upper-Bound Performance with Offline Variant.}

We also evaluate an offline variant of our model, which observes the full conditioning signal $\mathbf{c}_{[t-h:t+f]}$.
This offline version provides an upper-bound estimate for our architecture, while our standard online model remains competitive despite strictly satisfying causal online constraints.

\subsection{Noise Robust Reconstruction}
\label{sec:eval_robust}

Next, we evaluate the robustness of our method under noisy egocentric observations in Table~\ref{tab:quantitative_robust}. 
Under strict online constraints, our approach steadily generates high-quality motions (e.g., Semantic Score of 0.971) while maintaining low reconstruction error (MPJPE of 0.084m) despite noisy egocentric signals. 
These results demonstrate that our noise-robust causal diffusion framework, trained with observation corruption, achieves strong robustness while maintaining high reconstruction accuracy, motion quality, and real-time feasibility.

\vspace{-0.5em}
\begin{table*}[h!]
\caption{Quantitative egocentric motion reconstruction results on the noisy EE4D-Motion dataset.}
    \centering
    \setlength{\tabcolsep}{6pt}
    \renewcommand{\arraystretch}{1.1}
    \scalebox{0.73}{
    \begin{tabular}{lcccccccc}
    \toprule
    \multirow{2}{*}{Methods}
    & \multicolumn{4}{c}{Reconstruction $\downarrow$}
    & \multicolumn{2}{c}{Quality}
    & \multicolumn{2}{c}{Smoothness}\\
    \cmidrule(lr){2-5}
    \cmidrule(lr){6-7}
    \cmidrule(lr){8-9}
     & Head PE & Wrist PE & MPJPE & MPJRE-F & Semantic $\uparrow$ & FID $\downarrow$ & PJ $\rightarrow$ & AUJ $\downarrow$ \\
    \midrule
    Real motions  & - & - & - & - & 0.999 & 0.001 & 0.329 & 10.7    \\
    \midrule
    AvatarJLM~\cite{avatarjlm} & - & 0.228 & 0.171 & 0.517 & 0.813 & 0.118 & 1.882 & 20.8   \\
    RPM~\cite{barquero2025rolling}  & - & 0.147 & 0.134 & 0.386 & 0.899 & 0.054 & 0.436 & 12.1 \\
    HMD$^2$$^\dagger$~\cite{guzov_jiang2025hmd2} & - & 0.149 & 0.155 & 0.283 & 0.884 & 0.046 & 0.391 & 11.3    \\
    EgoAllo~\cite{yi2025egoallo} & - & 0.091 & 0.152 & 0.635 & 0.846 & 0.048 & 0.937 & 15.3   \\
    UniEgoMotion~\cite{patel2025uniegomotion} & \textbf{0.053} & 0.134 & 0.145 & 0.523 & 0.902 & 0.030 & 0.874 & 14.7  \\
    \method\;(Ours)  & 0.081 & \textbf{0.086} & \textbf{0.084} & \textbf{0.163} & \textbf{0.971} & \textbf{0.021} & \textbf{0.323} & \textbf{11.0} \\
    \midrule
    \;\; w/o noise robust.  & 0.093 & 0.101 & 0.102 & 0.201 & 0.943 & 0.027 & 0.412 & 11.3  \\
    
    \bottomrule
    \end{tabular}
    }
    \label{tab:quantitative_robust}
\end{table*}

\vspace{-1em}
\begin{figure*}[h]
\begin{center}
\includegraphics[width=0.9\linewidth]{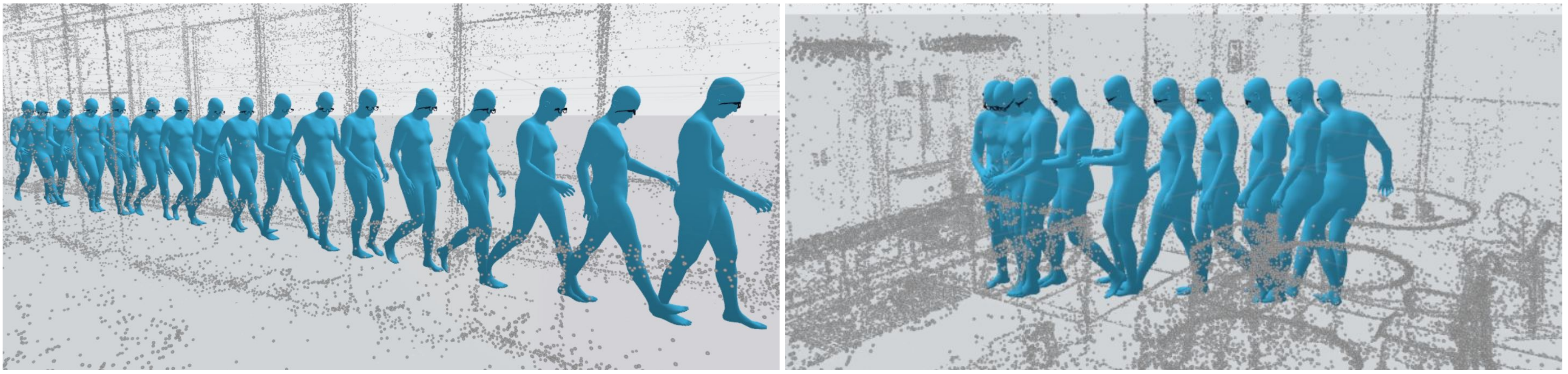}
\vspace{-0.5em}
\caption{
      Qualitative Ego-Exo4D examples using Project Aria SLAM trajectories and HaMeR hand estimates as streaming egocentric inputs. Despite real sensing noise and domain shift from training, our framework produces temporally coherent and plausible full-body motion.
}
\label{fig:egoexo4d}
\end{center}
\vspace{-1em}
\end{figure*}

To further assess robustness beyond EE4D-Motion, we additionally train our model on the AMASS dataset~\cite{AMASS:ICCV:2019} with synthetic egocentric. 
We provide additional results on the AMASS dataset in the Appendix.
We then apply the trained model to real-world Ego-Exo4D~\cite{grauman2024egoexo4d} sequences using actual SLAM trajectories and hand estimation~\cite{hamer} results from streaming egocentric inputs.
Our method produces stable and plausible motion reconstructions in these real-world settings, highlighting the framework's effectiveness in practice and its consistent performance across diverse data sources. 
\section{Conclusion}
\label{sec:5_conclusion}

In this paper, we presented EgoForce, a novel diffusion-based framework for online egocentric motion reconstruction.
By formulating egocentric reconstruction as a causal generation problem with temporally evolving uncertainty, we successfully adapted the expressive power of diffusion models to an online streaming setting.
Experimental results demonstrate that EgoForce generates high-quality motion from streaming egocentric observations, preserves long-term motion connectivity, and outperforms prior offline diffusion approaches, achieving state-of-the-art performance.
We expect EgoForce to promote the development of practical, real-world interactive applications in which continuous and reliable human motion understanding is essential for seamless daily-life assistance and interaction.

{

\bibliographystyle{plain}
\bibliography{main}
}

\end{document}